\documentclass[a4paper,fleqn]{cas-sc}

\usepackage[authoryear,longnamesfirst]{natbib}

\begin{document}
\let\WriteBookmarks\relax
\def\floatpagepagefraction{1}
\def\textpagefraction{.001}

\shorttitle{Dual-axis attribution of zebrafish tectal microcircuits}

\shortauthors{Li}

\title[mode=title]{Dual-axis attribution of zebrafish tectal microcircuits for energy-efficient and robust neurocomputing}

\author[1]{Ningping Li}
\ead{ningpingli@mail.ustc.edu.cn}
\credit{Conceptualization, Methodology, Software, Validation, Formal analysis, Writing -- original draft}

\author[1]{Hao Zhang}
\cormark[1]
\ead{hzhangxyz@ustc.edu.cn}
\credit{Supervision, Funding acquisition, Project administration, Writing -- review and editing}

\author[1]{Yi Zhou}
\cormark[1]
\ead{yi_zhou@ustc.edu.cn}
\credit{Supervision, Funding acquisition, Project administration, Writing -- review and editing}

\affiliation[1]{
  organization={University of Science and Technology of China},
  city={Hefei},
  country={China}
}

\cortext[1]{Corresponding author.}

\begin{abstract}
Biological neural circuits contain specialized substructures that support distinct computational functions, yet many bio-inspired neural networks borrow biological motifs without identifying their circuit-level origins.
In this study, we investigate whether zebrafish tectal microcircuits can be attributed along two computational axes: energy-efficient information processing and robustness-preserving stabilization.
We reconstruct a directed zebrafish-inspired retinotectal microcircuit graph and verify retinotectal signal propagation through dynamic simulation.
A leaky integrate-and-fire spiking neural network is then used as a nonlinear perturbation testbed, where predefined subcircuits are selectively ablated and evaluated using the Energy Sensitivity Index and the Robustness Sensitivity Index.
The results reveal a functional dissociation between two tectal subcircuits.
The \textit{ns\_TIN} subcircuit shows a low spike footprint but a measurable influence on prediction error, suggesting a role as a spike-efficient internal information gate.
In contrast, the \textit{superficial\_TIN} subcircuit produces the highest robustness sensitivity, suggesting a feedback-like role in maintaining system-level stability.
We further transfer these attributed functions into ResNet18-based artificial neural networks and evaluate them on CIFAR-10 under inference-budget reduction and Gaussian noise corruption.
The \textit{ns\_TIN}-inspired module improves performance preservation under reduced computation, whereas the \textit{superficial\_TIN}-inspired module improves robustness under input noise.
These findings provide a subcircuit-level route for linking biological circuit organization with bio-inspired neural architecture design.
\end{abstract}

\begin{highlights}
\item A zebrafish-inspired retinotectal microcircuit graph is reconstructed for subcircuit-level attribution.
\item A dual-axis framework distinguishes energy-sensitive and robustness-sensitive subcircuits.
\item \textit{ns\_TIN} and \textit{superficial\_TIN} show functionally dissociated computational roles.
\item The attributed functions are transferred to ResNet18-based models and validated on CIFAR-10.
\end{highlights}


\maketitle

\section{Introduction}
\label{sec:introduction}

Biological neural circuits are composed of specialized substructures rather than homogeneous computational substrates. 
These substructures support perception, integration, recurrent modulation, and sensorimotor transformation through distributed circuit motifs. 
A key question in neurocomputing is therefore not only whether biological principles can inspire artificial neural networks, but also which biological substructures implement which computational functions \cite{olshausen1996emergence,maass1997networks,yamins2016using,jonas2017could}.

The zebrafish brain provides a compact model for studying this question because whole-brain imaging, atlas registration, and cell-type-resolved circuit descriptions can be linked across scales. 
In the visual--motor system, visual signals from retinal ganglion cells are transmitted to the optic tectum, where tectal interneurons and projection neurons participate in visual filtering, local integration, and downstream motor-related transformation \cite{niell2005functional,delbene2010filtering,baier2024visual}. 
Because this system contains identifiable neuronal populations and atlas-aligned projection classes, including RGCs, tectal interneurons, tectal projection neurons, and downstream motor-related output units, it is well suited for building a mesoscopic--microscopic abstraction and examining whether distinct substructures play distinct computational roles \cite{gabriel2012layerspecific,kunst2019cellularresolution}.

Existing bio-inspired neural networks often borrow high-level motifs such as sparse coding, feedback, attention, gating, inhibition, or spiking dynamics \cite{hochreiter1997long,roy2019spikebased}. 
However, these motifs are usually transferred into artificial models without first determining their circuit-level biological origins. 
As a result, the link between biological organization and artificial neural architecture design often remains qualitative. 
A more mechanistic approach should identify functionally specific substructures, quantify their contributions through controlled perturbation, and test whether the attributed functions can be transferred to artificial architectures \cite{bau2017network,jonas2017could}.

In this study, we investigate the model-based functional specificity of zebrafish brain substructures in a visual--motor abstraction. 
Rather than treating the retinotectal system as a single inspiration source, we ask whether a two-level brain model, with microscopic cell-category nodes and mesoscopic substructure groups, can be used to identify specific structures that are sensitive to energy cost and robustness degradation. 
Specifically, we test whether some substructures behave as energy-efficient information gates, while others behave as robustness-efficient stabilizing structures. 
To this end, we construct a zebrafish visual--motor brain graph, perform substructure-level ablation in a spiking neural network, and evaluate whether selected functions can be transferred to ResNet18-based artificial neural networks \cite{neftci2019surrogate,he2016deep}.

We propose a dual-axis attribution framework for substructure analysis. 
The Energy Sensitivity Index (ESI) measures task-performance contribution relative to spike activity, thereby identifying energy-efficient information-processing structures. 
The Robustness Sensitivity Index (RSI) measures normalized performance degradation after substructure removal, thereby identifying substructures whose ablation strongly disrupts the model. 
This formulation separates efficient information routing from dynamic stabilization, rather than treating substructure importance as a single scalar value.

Our results identify two specific sensitive substructures from the zebrafish-inspired abstraction. 
\textit{ns\_TIN} has a low energy footprint, estimated by spike count, but a measurable effect on prediction error, suggesting an energy-efficient bottleneck role. 
\textit{superficial\_TIN} shows the strongest robustness sensitivity, suggesting a candidate stabilizing role. 
These findings indicate that zebrafish brain substructures differ not only in overall importance but also in computational specialization.

We further test whether these functions remain meaningful outside the original SNN. 
The \textit{ns\_TIN}-inspired module is evaluated under computation-budget reduction, while the \textit{superficial\_TIN}-inspired module is evaluated under Gaussian noise corruption, a common setting for robustness evaluation \cite{hendrycks2019benchmarking,spoerer2017recurrent}. 
These transfer experiments do not aim to claim universal architectural superiority, but to examine whether biologically attributed functions can guide ANN module design.

The main contributions are summarized as follows:

\begin{enumerate}
  \item We construct a $52 \times 52$ zebrafish visual--motor brain abstraction based on biological connection probabilities and organize it at two levels: microscopic cell-category nodes and mesoscopic functional substructures.

  \item We propose a dual-axis substructure attribution framework using ESI for energy-efficient information-processing structures and RSI for robustness-efficient structures.

  \item We identify two functionally specific sensitive substructures: \textit{ns\_TIN} for energy-efficient bottleneck gating and \textit{superficial\_TIN} for robustness-efficient stabilization.

  \item We transfer two representative attributed substructures into ResNet18-based ANN modules and evaluate their functional specificity on CIFAR-10 under computation-budget reduction and Gaussian noise corruption.
\end{enumerate}

\section{Related work}
\label{sec:related_work}

\subsection{Bio-inspired neural computation, spiking models, and robust artificial networks}
\label{subsec:bio_snn_robust}

Biological neural systems have long inspired artificial neural networks. 
Sparse coding showed that efficient representations learned from natural images can resemble receptive-field properties in visual cortex \cite{olshausen1996emergence}, while recurrent, gated, and attention-based mechanisms have influenced models for temporal integration, memory, and dynamic information routing \cite{hochreiter1997long}. 
Spiking neural networks further approximate biological computation by using discrete spike events instead of continuous activations \cite{maass1997networks}. 
Because spike-based computation is sparse and event-driven, SNNs are widely studied for energy-efficient neural computation \cite{roy2019spikebased,tavanaei2019deep,pfeiffer2018deep}. 
Recent ANN-to-SNN conversion, surrogate-gradient learning, and neuromorphic hardware have further improved the practicality of SNNs \cite{rueckauer2017conversion,neftci2019surrogate,merolla2014million,davies2018loihi}. 
However, low activity cost does not necessarily indicate functional importance. 
A substructure may be sparse but irrelevant, or active but redundant. 
Therefore, this study evaluates not only energy footprint, estimated from spike activity, but also the performance contribution relative to this activity cost.

Robustness is another central issue in neural computation. 
Artificial neural networks can be sensitive to adversarial perturbations, common corruptions, structural ablation, pruning, and noisy inputs \cite{szegedy2013intriguing,goodfellow2014explaining,hendrycks2019benchmarking}. 
Studies on shape bias, feature denoising, and recurrent processing suggest that architectural mechanisms can improve stability under distribution shift or degraded sensory input \cite{geirhos2018imagenettrained,xie2019feature,spoerer2017recurrent}. 
Nevertheless, most robustness studies focus on whole-model behavior rather than identifying which specific biological or bio-inspired substructures preserve stability. 
Similarly, many bio-inspired models adopt sparse coding, feedback, inhibition, or gating as abstract motifs, but do not first determine their circuit-level origins. 
Goal-driven modeling and network interpretability have connected artificial networks with neural data and internal functional units \cite{yamins2016using,bau2017network}, while lesion-style analysis highlights the need for perturbation-based functional testing \cite{jonas2017could}. 
Our work follows this direction by combining biological brain-atlas priors, SNN substructure ablation, and ANN transfer evaluation.

\subsection{Zebrafish visual--motor circuits and substructure-level functional attribution}
\label{subsec:zebrafish_attribution}

The larval zebrafish visual system is a compact and accessible model for studying vertebrate sensorimotor computation \cite{baier2024visual}. 
Visual signals from retinal ganglion cells are transmitted to the optic tectum, which serves as a major hub for transforming visual input into orienting, prey-capture, avoidance, and other motor-related behaviors \cite{niell2005functional,baier2024visual}. 
This makes the retinotectal pathway suitable for analyzing how local circuit structures support perception, integration, and sensorimotor transformation.

Previous studies show that the zebrafish optic tectum is not a homogeneous visual-processing region. 
Retinal inputs are organized into laminar and feature-selective pathways, while tectal interneurons contribute to filtering, motion detection, local modulation, and information routing \cite{delbene2010filtering,gabriel2012layerspecific,robles2011characterization,yin2019optic}. 
Anatomical and genetic studies have further identified diverse tectal neuron classes with distinct morphologies, neurotransmitter phenotypes, and projection patterns \cite{demarco2020neuron,kunst2019cellularresolution}. 
At the behavioral level, tectal and downstream circuits are involved in prey capture, prey detection, pursuit, and approach--avoidance selection \cite{gahtan2005visual,bianco2011prey,semmelhack2014dedicated,forster2020retinotectal,barker2015sensorimotor}. 
Recent modeling work also suggests that tectal circuits may support reservoir-like visuomotor transformation \cite{qian2025adaptive}. 
Together, these studies indicate that tectal computation is distributed across specialized substructures.

However, existing zebrafish studies mainly characterize biological structure, neural responses, or behavior, while artificial bio-inspired models often borrow motifs such as feedback, gating, inhibition, or sparse activity without identifying the responsible biological substructures. 
This leaves a gap between biological circuit organization and artificial neural architecture design. 
Substructure-level perturbation offers a way to bridge this gap, but naive lesion analysis can be misleading if it only measures global performance loss \cite{jonas2017could}. 
Therefore, functional attribution should distinguish different computational axes rather than treating importance as a single scalar value.

In this study, we construct a zebrafish visual--motor brain abstraction and perform substructure-level ablation in an SNN. 
We use the Energy Sensitivity Index to identify substructures whose performance contribution is disproportionate to their energy footprint, and the Robustness Sensitivity Index to identify substructures whose removal causes strong system-level degradation. 
Finally, we transfer two representative attributed substructures into ResNet18-based ANN modules and evaluate them under matched conditions: computation-budget reduction for the energy-efficient prototype and Gaussian noise corruption for the robustness-efficient prototype.

\section{Materials and methods}
\label{sec:methods}

This study aims to build a mesoscopic--microscopic abstraction of zebrafish brain circuitry and computationally attribute functionally specific substructures that are sensitive to energy cost and robustness degradation. 
The proposed framework consists of five stages: biological graph construction, mesoscopic substructure definition, dynamic feasibility checking, SNN-based substructure ablation, and ANN transfer.

First, biological connection priors from the zebrafish visual--motor system are used to construct a directed brain abstraction. 
Second, microscopic cell-category nodes are grouped into mesoscopic anatomical and functional substructures. 
Third, dynamic simulation is used to test whether the reconstructed graph supports retinotectal signal propagation. 
Fourth, a spiking neural network is used as a nonlinear testbed for substructure-level perturbation and attribution. 
Finally, selected attributed substructures are transferred into artificial neural network modules to examine whether their computational roles remain meaningful outside the original SNN setting.

The central methodological principle is that biological substructures should first be screened through controlled perturbation before their roles are transferred into artificial architectures.

\subsection{Mesoscopic--microscopic zebrafish brain abstraction}
\label{subsec:graph_and_simulation}

We reconstructed a directed zebrafish visual--motor brain abstraction based on anatomical and electrophysiological connection priors. 
Each node represents a neural category or subpopulation rather than an individual neuron, and each directed edge represents a non-zero connection probability from a presynaptic category to a postsynaptic category. 
This microscopic level preserves major neuronal-class distinctions while allowing the circuit to be represented as a structured connection matrix.
The implementation uses the complete node list and connection-probability table stored in the accompanying code repository as \texttt{code/data/connection\_matrix\_complete.json}; the same file records 52 node categories, 938 non-zero directed entries, a density of 0.3469, and connection probabilities ranging from 0.000271 to 0.285035.

The connection matrix is denoted as
\begin{equation}
  A \in \mathbb{R}^{52 \times 52},
  \label{eq:connection_matrix}
\end{equation}
where $A_{ij}$ represents the connection probability from presynaptic category $j$ to postsynaptic category $i$. 
Thus, $A_{ij}>0$ indicates a directed projection from node $j$ to node $i$. 
Rows of the matrix correspond to postsynaptic categories and columns correspond to presynaptic categories. 
The major microscopic node groups include retinal ganglion cells (RGCs), tectal interneurons (TINs), tectal projection neurons (TPNs), downstream motor-related units, and a task-specific readout category.

To characterize the global topology of the reconstructed graph, we used connection density and spectral radius. 
The connection density is defined as
\begin{equation}
  D = \frac{\left|\{A_{ij} > 0\}\right|}{N^2},
  \label{eq:connection_density}
\end{equation}
where $N=52$. 
The spectral radius is defined as
\begin{equation}
  \rho(A) = \max_i |\lambda_i(A)|,
  \label{eq:spectral_radius}
\end{equation}
where $\lambda_i(A)$ denotes the $i$-th eigenvalue of $A$.

\begin{table}[t]
  \centering
  \caption{Composition of microscopic categories in the zebrafish visual--motor brain abstraction.}
  \label{tab:node_composition}
  \begin{tabular}{lll}
    \toprule
    Category & Representative nodes & Functional role \\
    \midrule
    RGC & SO-A RGC, SGC/SAC-P RGC & Visual input \\
    TIN & \textit{ns\_TIN}, \textit{superficial\_TIN}, \textit{deep\_TIN} & Local integration and inhibition \\
    TPN & TPN-O, TPN-E & Tectal projection output \\
    Integration/output & SIN, TPN-E, TPN-O & Downstream integration and projection output \\
    \bottomrule
  \end{tabular}
\end{table}

At the mesoscopic level, microscopic nodes are grouped into anatomically interpretable or computationally motivated substructures. 
The predefined groups include input/output structures (\textit{RGC\_input}, \textit{TPN\_output}), excitation-defined TIN populations (\textit{excitatory\_TIN}, \textit{inhibitory\_TIN}), layer-defined groups (\textit{S12\_group}, \textit{S34\_group}, \textit{S56\_group}, \textit{superficial\_TIN}, \textit{deep\_TIN}), morphology-related groups (\textit{SGC\_group}, \textit{SAC\_group}), and integration hubs (\textit{SIN\_hub}, \textit{TPN\_hub}, \textit{integration\_hubs}). 
In addition, agglomerative clustering of node-level topological features provides data-driven mesoscopic communities that complement the hand-defined biological groups.

\begin{table}[t]
  \centering
  \caption{Two-level abstraction used for zebrafish brain substructure attribution.}
  \label{tab:meso_micro_abstraction}
  \begin{tabular}{lll}
    \toprule
    Level & Representation & Examples \\
    \midrule
    Microscopic & 52 cell-category nodes & \textit{e\_ns\_TIN}, \textit{i\_ns\_TIN}, TPN-E, TPN-O \\
    Mesoscopic & Anatomical or functional substructures & \textit{ns\_TIN}, \textit{S12\_group}, \textit{superficial\_TIN} \\
    Data-driven mesoscopic & Topology-derived communities & Six agglomerative clusters from connection features \\
    \bottomrule
  \end{tabular}
\end{table}

\begin{figure}[t]
  \centering
  \includegraphics[width=\linewidth]{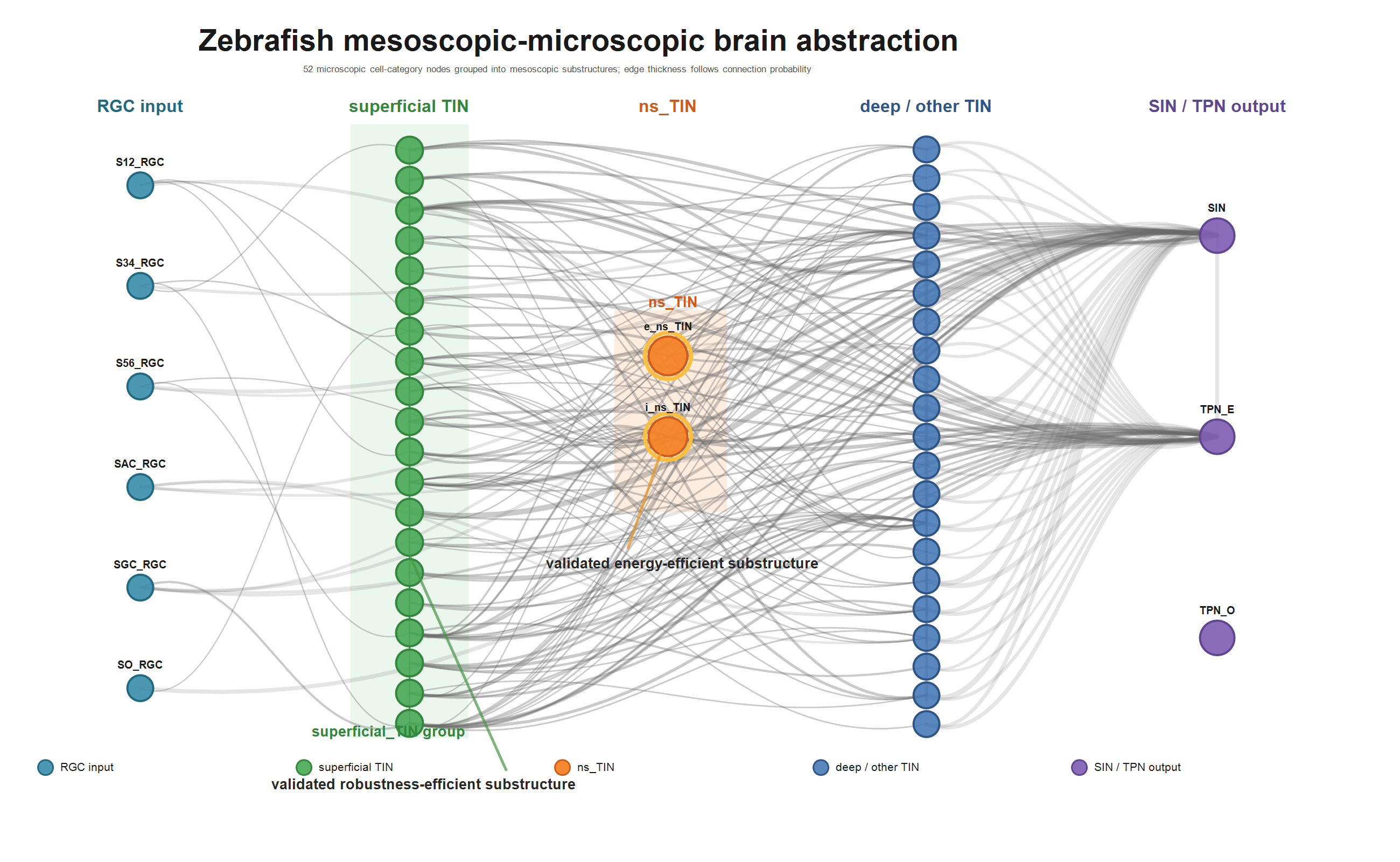}
  \caption{Mesoscopic--microscopic abstraction of the zebrafish visual--motor brain graph. Microscopic cell-category nodes are arranged by functional group, and gray edges denote directed connection probabilities. The \textit{ns\_TIN} group is highlighted as the validated energy-efficient substructure, whereas the \textit{superficial\_TIN} group is highlighted as the validated robustness-efficient substructure.}
  \label{fig:micro_meso_abstraction}
\end{figure}

The two substructures selected for cross-modal transfer are shown topologically in Fig.~\ref{fig:micro_meso_abstraction}. 
This visualization is not used as additional experimental evidence; instead, it documents how microscopic cell-category nodes are organized into mesoscopic substructures and how the two identified substructures are embedded in the reconstructed connection graph before perturbation analysis.

After graph reconstruction, dynamic simulation was used to test whether the graph can support signal propagation along retinotectal pathways. 
This stage served as a feasibility check rather than as the main performance evaluation. 
RGC populations were treated as visual input nodes, and signals were propagated through intermediate TIN populations to TPN and motor-related output populations.

Exponential synapses were used to model asynchronous synaptic transmission. 
Following the convention that $A_{ij}$ denotes a projection from presynaptic category $j$ to postsynaptic category $i$, the synaptic current from $j$ to $i$ is written as
\begin{equation}
  \tau_s \frac{d I_{ij}(t)}{dt}
  =
  - I_{ij}(t) + w_{ij} S_j(t),
  \label{eq:exp_synapse}
\end{equation}
where $I_{ij}(t)$ is the synaptic current received by category $i$, $\tau_s$ is the synaptic time constant, $w_{ij}$ is derived from $A_{ij}$, and $S_j(t)$ is the presynaptic spike or activity signal.

Two retinotectal propagation pathways were considered: the RGC-to-TPN-O pathway, denoted as R2O, and the RGC-to-TPN-E pathway, denoted as R2E. 
The corresponding simulation settings and feasibility-check results are reported in the experimental section.

\subsection{SNN perturbation model and dual-axis attribution}
\label{subsec:snn_attribution}

To quantify the computational contribution of zebrafish-inspired substructures, we used a leaky integrate-and-fire SNN as a nonlinear perturbation testbed. 
The SNN was evaluated on Lorenz chaotic trajectory prediction because the Lorenz system provides a nonlinear and dynamically unstable environment. 
The task was used for comparative substructure attribution rather than for demonstrating state-of-the-art chaotic forecasting performance.

The standard Lorenz system is defined as
\begin{equation}
\begin{aligned}
  \frac{dx}{dt} &= \sigma (y - x), \\
  \frac{dy}{dt} &= x(\rho - z) - y, \\
  \frac{dz}{dt} &= xy - \beta z,
\end{aligned}
\label{eq:lorenz}
\end{equation}
where $x$, $y$, and $z$ are the system states, and $\sigma$, $\rho$, and $\beta$ are system parameters.

The SNN uses leaky integrate-and-fire neurons. 
For neuron $i$, the membrane potential follows
\begin{equation}
  \tau_m \frac{dV_i(t)}{dt}
  =
  -\left(V_i(t) - V_{\mathrm{rest}}\right) + R_m I_i(t),
  \label{eq:lif_voltage}
\end{equation}
where $V_i(t)$ is the membrane potential, $\tau_m$ is the membrane time constant, $V_{\mathrm{rest}}$ is the resting potential, $R_m$ is the membrane resistance, and $I_i(t)$ is the input current. 
A spike is generated when the membrane potential reaches the firing threshold:
\begin{equation}
  S_i(t) =
  \begin{cases}
    1, & V_i(t) \geq V_{\mathrm{th}}, \\
    0, & V_i(t) < V_{\mathrm{th}}.
  \end{cases}
  \label{eq:spike_generation}
\end{equation}

The SNN uses the reconstructed 52-category computational graph. 
Model training details, Lorenz sequence-generation settings, and baseline performance are reported in the experimental section.

To examine functional specificity, we performed substructure-level ablation in the trained SNN. 
Each substructure is defined as a set of neural categories sharing anatomical or functional identity, such as RGC input groups, TIN subgroups, TPN output groups, or motor-related groups. 
For a target substructure $\mathcal{G}$, ablation is implemented by disabling the corresponding neural categories and their associated synaptic projections during evaluation.

Let $\mathcal{V}$ denote the full set of neural categories and $\mathcal{G} \subset \mathcal{V}$ denote the ablated substructure. 
The ablated connection matrix $A^{(-\mathcal{G})}$ is defined as
\begin{equation}
  A^{(-\mathcal{G})}_{ij}
  =
  \begin{cases}
    0, & i \in \mathcal{G} \ \mathrm{or}\ j \in \mathcal{G}, \\
    A_{ij}, & \mathrm{otherwise}.
  \end{cases}
  \label{eq:ablation_matrix}
\end{equation}
This operation removes both incoming and outgoing interactions associated with the target substructure. 
The model is evaluated without further training, so that performance changes can be attributed to the removed substructure rather than to compensatory learning.

For each ablation condition, we measure post-ablation MSE and spike count. 
Changes relative to the intact baseline model are used to compute the Energy Sensitivity Index (ESI) and Robustness Sensitivity Index (RSI). 
The ESI is defined as
\begin{equation}
  \mathrm{ESI} =
  \frac{\Delta \mathrm{Spike}\%}
  {|\Delta \mathrm{MSE}\%| + \epsilon},
  \label{eq:esi}
\end{equation}
where $\epsilon$ is a small positive constant used to avoid division by zero. 
The relative spike change is defined as
\begin{equation}
  \Delta \mathrm{Spike}\% =
  \frac{
  \mathrm{Spikes}_{\mathrm{after}} - \mathrm{Spikes}_{\mathrm{before}}
  }
  {
  \mathrm{Spikes}_{\mathrm{before}}
  }
  \times 100\%,
  \label{eq:delta_spike}
\end{equation}
and the relative MSE change is defined as
\begin{equation}
  \Delta \mathrm{MSE}\% =
  \frac{
  \mathrm{MSE}_{\mathrm{after}} - \mathrm{MSE}_{\mathrm{before}}
  }
  {
  \mathrm{MSE}_{\mathrm{before}}
  }
  \times 100\%.
  \label{eq:delta_mse_percent}
\end{equation}

Spike count is used as a proxy for neural activity cost rather than a direct measurement of physical energy consumption. 
For ablations that increase prediction error and reduce spike count, the negative sign of ESI reflects the direction of the activity-cost change. 
Substructures with small $|\mathrm{ESI}|$ values cause relatively large error increases for a small energy-cost reduction, estimated by spike count, and are therefore interpreted as energy-efficient information-processing candidates. 
Direct visual input groups are analyzed separately from internal processing substructures.

The RSI is defined as
\begin{equation}
  \mathrm{RSI} =
  \frac{\Delta \mathrm{MSE}}
  {\mathrm{MSE}_{\mathrm{baseline}}},
  \label{eq:rsi}
\end{equation}
where
\begin{equation}
  \Delta \mathrm{MSE}
  =
  \mathrm{MSE}_{\mathrm{after}}
  -
  \mathrm{MSE}_{\mathrm{baseline}}.
  \label{eq:delta_mse}
\end{equation}
A higher RSI indicates that removing the corresponding substructure causes stronger relative performance degradation. 
RSI is therefore used as an ablation-based robustness-sensitivity score and as a proxy for identifying robustness-efficient substructures that may be important for maintaining system-level stability.

\subsection{Cross-modal transfer design}
\label{subsec:ann_transfer_methods}

To examine whether the attributed biological functions can be transferred beyond the SNN setting, we designed two ResNet18-based ANN variants. 
The goal is not to demonstrate universal architectural superiority, but to test whether functions identified through biological-SNN attribution remain meaningful in a conventional ANN setting.

\begin{figure}[htbp]
  \centering
  \includegraphics[width=0.8\linewidth]{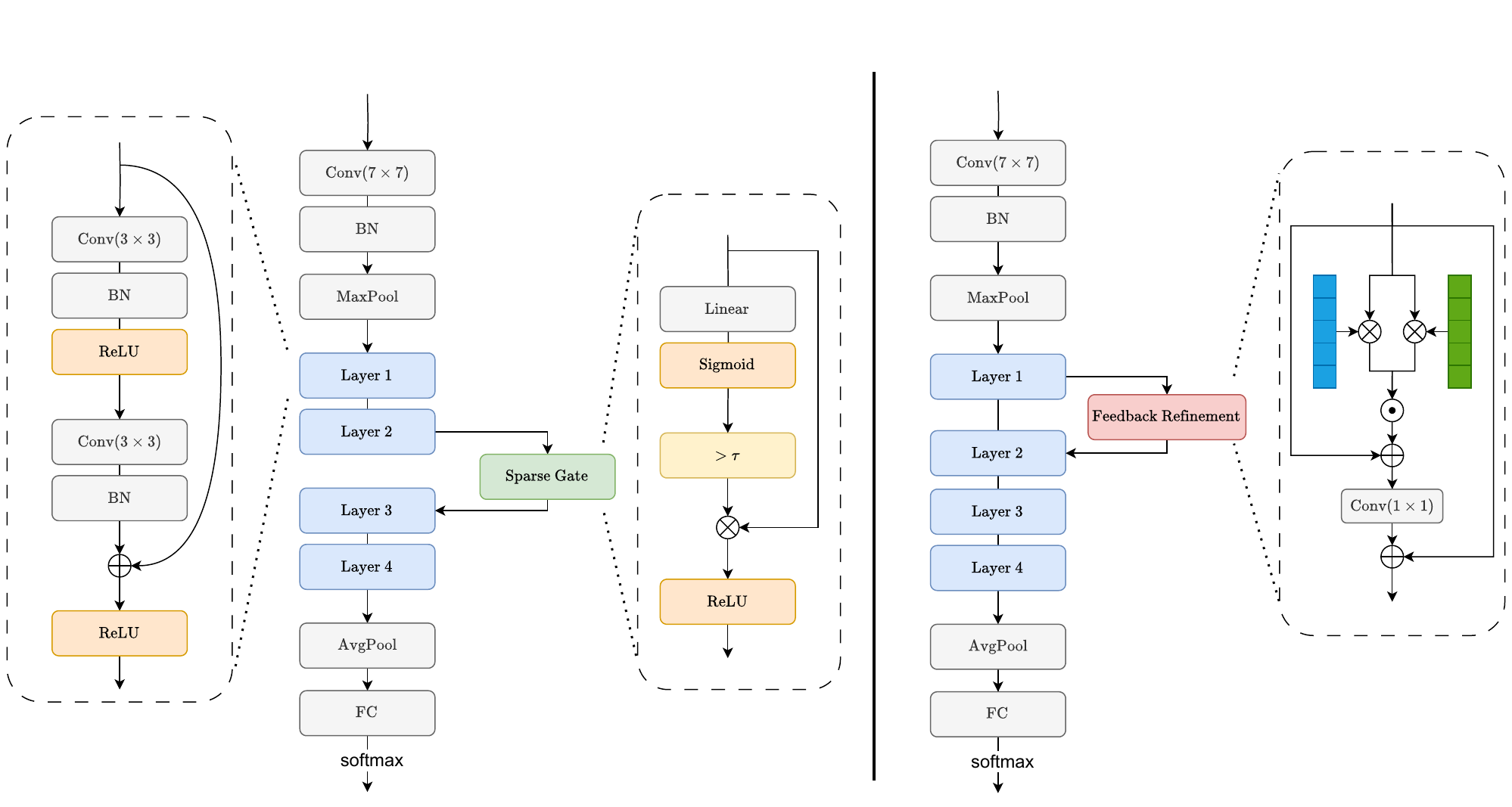}
  \caption{Schematic illustration of the ANN transfer design. The \textit{ns\_TIN}-inspired module introduces adaptive gating for computation-budget reduction, whereas the \textit{superficial\_TIN}-inspired module introduces feedback-like refinement for robustness under Gaussian noise corruption.}
  \label{fig:ann_transfer}
\end{figure}

The first variant, \textit{ResNet18WithNsTIN}, introduces an \textit{ns\_TIN}-inspired adaptive gating module after layer 2 of a CIFAR-style ResNet18 backbone. 
The module globally pools the 128-channel feature map, maps the pooled vector to two sigmoid gate values, thresholds the gates at 0.5 to obtain a sparse two-unit activation, projects it back to 128 channels, and applies the resulting channel-wise modulation through a residual multiplication. 
This module is designed to test the energy-efficient gating role attributed to \textit{ns\_TIN}. 
The second variant, \textit{ResNet18WithSuperficialTIN}, introduces a \textit{superficial\_TIN}-inspired feedback-like module after layer 1. 
This module uses a $1 \times 1$ projection from 64 channels to a 64-channel hidden state, applies a gated recurrent interaction over spatial positions, normalizes the hidden state with layer normalization, projects it back to the input channel dimension, and adds it to the residual feature map with a fixed scaling factor of 0.3. 
This module is designed to test the robustness-efficient stabilization role attributed to \textit{superficial\_TIN}. 
In the inference-budget experiments, budget control is implemented by residual-branch skipping over the four residual blocks in layer 3 and layer 4; in the noise experiments, Gaussian noise is added in pixel space before normalization.
The overall transfer design is shown in Fig.~\ref{fig:ann_transfer}.

\begin{table}[t]
  \centering
  \caption{Bio-inspired ResNet18 variants used for cross-modal transfer.}
  \label{tab:ann_models}
  \begin{tabular}{lll}
    \toprule
    Model & Description & Biological source \\
    \midrule
    BaselineResNet18 & Standard ResNet18 & None \\
    \textit{ResNet18WithNsTIN} & Adaptive gating module & \textit{ns\_TIN} \\
    \textit{ResNet18WithSuperficialTIN} & Feedback-like refinement module & \textit{superficial\_TIN} \\
    \bottomrule
  \end{tabular}
\end{table}

\section{Results}
\label{sec:results}

This section reports the experimental results of the proposed substructure-centered attribution framework. 
We first summarize the experimental setup for SNN-based perturbation analysis and ANN transfer evaluation. 
We then report the structural statistics and dynamic feasibility-check results of the reconstructed zebrafish visual--motor brain abstraction. 
Finally, we present the SNN substructure attribution results, identify two specific sensitive substructures, and report the cross-modal ANN transfer results.

\subsection{Experimental setup}
\label{subsec:experimental_setup}

The SNN experiments were conducted on a Lorenz chaotic trajectory prediction task. 
The reconstructed zebrafish-inspired graph was used as the computational substrate, and substructure-level ablation was performed after training. 
The model was not retrained after ablation, so that performance changes could be attributed to the removed substructure rather than to compensatory learning. 
The SNN was trained for 60 epochs with a prediction horizon of 100 time steps. 
Model performance was evaluated using mean squared error (MSE), coefficient of determination ($R^2$), correlation, and average spike count per sample. 
For each ablation condition, the post-ablation MSE and spike count were recorded to compute the Energy Sensitivity Index (ESI) and Robustness Sensitivity Index (RSI).

\begin{table}[!htbp]
  \centering
  \caption{Experimental setup for SNN-based substructure attribution.}
  \label{tab:snn_experimental_setup}
  \begin{tabular}{lc}
    \toprule
    Parameter & Value \\
    \midrule
    Task & Lorenz trajectory prediction \\
    Graph substrate & Zebrafish-inspired visual--motor brain abstraction \\
    Matrix size & $52 \times 52$ \\
    Neuron model & Leaky integrate-and-fire \\
    Training epochs & 60 \\
    Prediction horizon & 100 time steps \\
    Ablation strategy & Disable target nodes and associated projections \\
    Retraining after ablation & No \\
    Evaluation metrics & MSE, $R^2$, correlation, spike count \\
    Attribution indices & ESI, RSI \\
    \bottomrule
  \end{tabular}
\end{table}

The ANN transfer experiments were conducted on CIFAR-10 using ResNet18 as the backbone. 
The dataset was divided into 40,000 training images, 10,000 validation images, and 10,000 test images. 
All models were trained with the same optimization configuration to ensure a fair comparison. 
Two bio-inspired ResNet18 variants were evaluated: \textit{ResNet18WithNsTIN}, which introduces an adaptive gating module motivated by \textit{ns\_TIN}, and \textit{ResNet18WithSuperficialTIN}, which introduces a feedback-like refinement module motivated by \textit{superficial\_TIN}.

\begin{table}[!htbp]
  \centering
  \caption{Bio-inspired ResNet18 variants used in the ANN transfer experiments.}
  \label{tab:ann_models_results}
  \begin{tabular}{lll}
    \toprule
    Model & Description & Biological source \\
    \midrule
    BaselineResNet18 & Standard ResNet18 & None \\
    \textit{ResNet18WithNsTIN} & Adaptive gating module & \textit{ns\_TIN} \\
    \textit{ResNet18WithSuperficialTIN} & Feedback-like refinement module & \textit{superficial\_TIN} \\
    \bottomrule
  \end{tabular}
\end{table}

\begin{table}[!htbp]
  \centering
  \caption{Experimental setup for ANN transfer on CIFAR-10.}
  \label{tab:ann_setup}
  \begin{tabular}{lc}
    \toprule
    Parameter & Value \\
    \midrule
    Dataset & CIFAR-10 \\
    Training set & 40,000 \\
    Validation set & 10,000 \\
    Test set & 10,000 \\
    Backbone & ResNet18 \\
    Epochs & 200 \\
    Batch size & 128 \\
    Optimizer & SGD \\
    Learning rate & 0.1 \\
    Momentum & 0.9 \\
    Weight decay & $5 \times 10^{-4}$ \\
    Scheduler & CosineAnnealingLR \\
    Loss function & Cross-entropy \\
    Early stopping patience & 10 \\
    Random seed & 42 \\
    \bottomrule
  \end{tabular}
\end{table}

For the \textit{ns\_TIN}-inspired model, inference-budget reduction was tested using budget ratios of $1.00$, $0.75$, $0.50$, and $0.25$. 
Following the accompanying implementation, reduced budgets were applied at inference time by progressively disabling residual branches in the four residual blocks of layer 3 and layer 4, while preserving the shortcut path.
The budget degradation score is defined as
\begin{equation}
  S_c =
  \frac{1}{n}
  \sum_{b_i < 1.0}
  \frac{\mathrm{Acc}(1.0) - \mathrm{Acc}(b_i)}
  {1.0 - b_i},
  \label{eq:sc}
\end{equation}
where $b_i$ denotes the inference budget ratio and $n$ is the number of reduced-budget settings. 
A smaller $S_c$ indicates slower accuracy degradation as available computation decreases.

For the \textit{superficial\_TIN}-inspired model, Gaussian-noise robustness was tested using noise levels of $\sigma = 0.00$, $0.02$, $0.04$, $0.06$, and $0.08$. 
Noise was added to pixel-space images before normalization and clipping.
The noise degradation score is defined as
\begin{equation}
  S_n =
  \frac{1}{m}
  \sum_{\sigma_j > 0}
  \frac{\mathrm{Acc}(0.0) - \mathrm{Acc}(\sigma_j)}
  {\sigma_j},
  \label{eq:sn}
\end{equation}
where $\sigma_j$ denotes the standard deviation of Gaussian noise and $m$ is the number of non-zero noise settings. 
A smaller $S_n$ indicates slower accuracy degradation as input noise increases.

\subsection{Structural statistics and dynamic feasibility of the brain abstraction}
\label{subsec:structural_dynamic_results}

The reconstructed zebrafish visual--motor brain abstraction is represented as a $52 \times 52$ directed connection probability matrix. 
Each matrix entry corresponds to the directed connection probability between two neural categories or subpopulations. 
The graph contains 938 non-zero directed connection types, corresponding to a density of 34.7\%. 
The spectral radius of the connection matrix is 1.5173.

These results indicate that the reconstructed graph is neither fully connected nor extremely sparse, but has a moderately dense and selective topology. 
The directed connection pattern is also non-uniform across retinal ganglion cells (RGCs), tectal interneurons (TINs), tectal projection neurons (TPNs), and downstream motor-related units. 
This non-uniform organization provides the structural basis for testing whether different substructures contribute to distinct computational functions.

\begin{figure}[!htbp]
  \centering
  \includegraphics[width=0.8\linewidth]{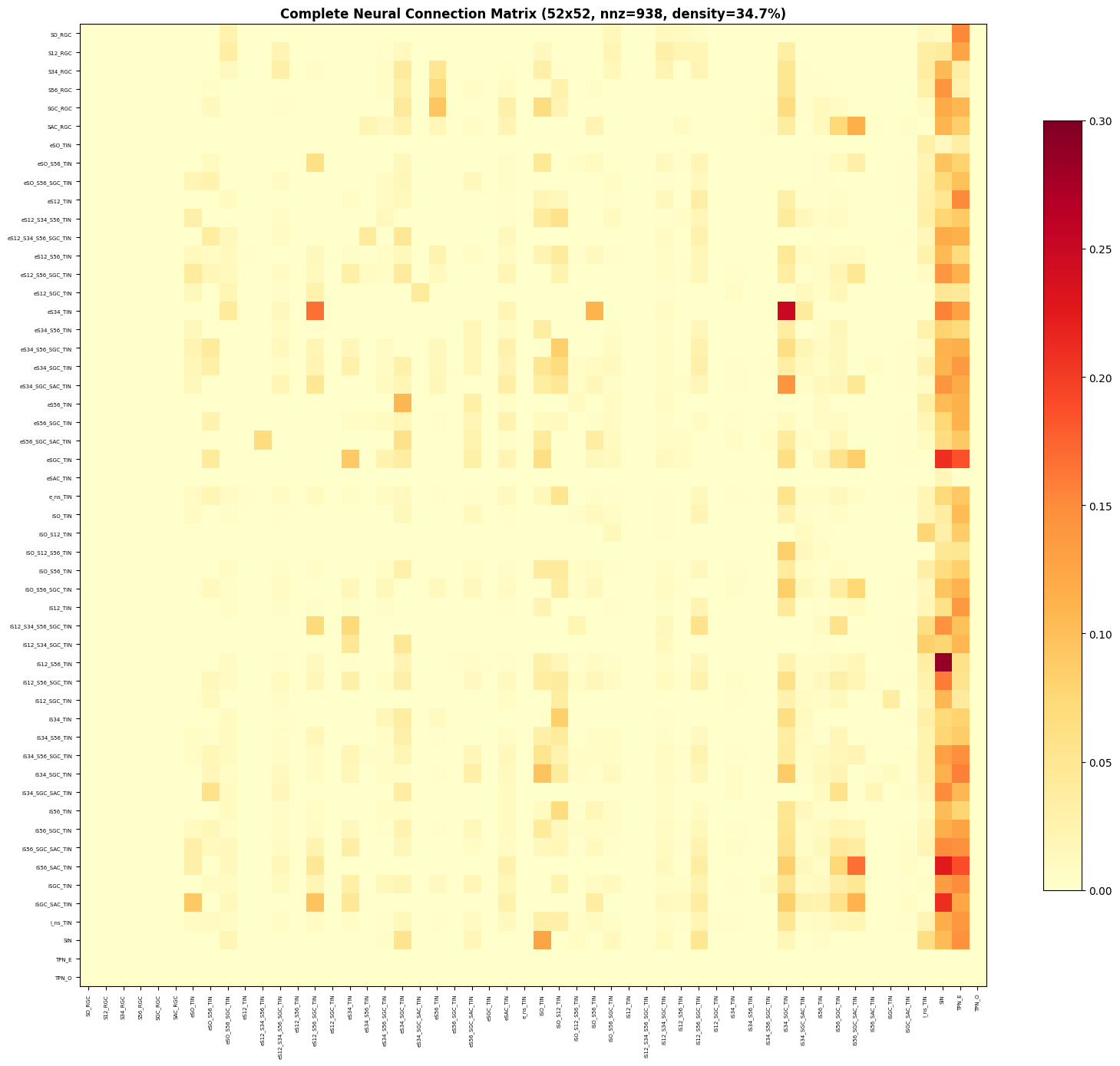}
  \caption{Directed connection matrix of the reconstructed zebrafish-inspired visual--motor brain abstraction. Rows indicate postsynaptic categories and columns indicate presynaptic categories. Nodes are grouped by neural class.}
  \label{fig:connection_matrix}
\end{figure}

\begin{table}[!htbp]
  \centering
  \caption{Graph-level statistics of the reconstructed zebrafish visual--motor brain abstraction.}
  \label{tab:graph_statistics}
  \begin{tabular}{lc}
    \toprule
    Metric & Value \\
    \midrule
    Matrix size & $52 \times 52$ \\
    Non-zero directed connections & 938 \\
    Connection density & 34.7\% \\
    Spectral radius & 1.5173 \\
    \bottomrule
  \end{tabular}
\end{table}

We further examined whether the reconstructed graph supports activity propagation along retinotectal pathways. 
BrainPy-based dynamic simulation showed that visual input signals from RGC populations could propagate through intermediate TIN populations and reach TPN and motor-related output populations. 
Both examined pathways, the RGC-to-TPN-O pathway and the RGC-to-TPN-E pathway, produced activity propagation from visual input nodes to downstream output-related nodes. 
This result supports the dynamic feasibility of the reconstructed abstraction and justifies its use as a computational substrate for subsequent SNN-based ablation analysis, but it should not be interpreted as a direct validation of the biological circuit.

\subsection{SNN-based substructure attribution}
\label{subsec:snn_substructure_attribution}

Before substructure ablation, we evaluated the intact SNN on the Lorenz chaotic trajectory prediction task. 
The baseline model achieved an MSE of 0.9318, an $R^2$ of 0.0751, a correlation of 0.2800, and an average spike count of 3349.09 per sample. 
The relatively low $R^2$ indicates that this experiment should be interpreted as a comparative perturbation analysis rather than as a demonstration of strong absolute forecasting performance.

\begin{table}[!htbp]
  \centering
  \caption{Baseline SNN performance on Lorenz chaotic trajectory prediction.}
  \label{tab:snn_baseline_results}
  \begin{tabular}{lc}
    \toprule
    Metric & Value \\
    \midrule
    MSE & 0.9318 \\
    $R^2$ & 0.0751 \\
    Correlation & 0.2800 \\
    Average spikes/sample & 3349.09 \\
    \bottomrule
  \end{tabular}
\end{table}

\begin{figure}[!htbp]
  \centering
  \includegraphics[width=0.95\linewidth]{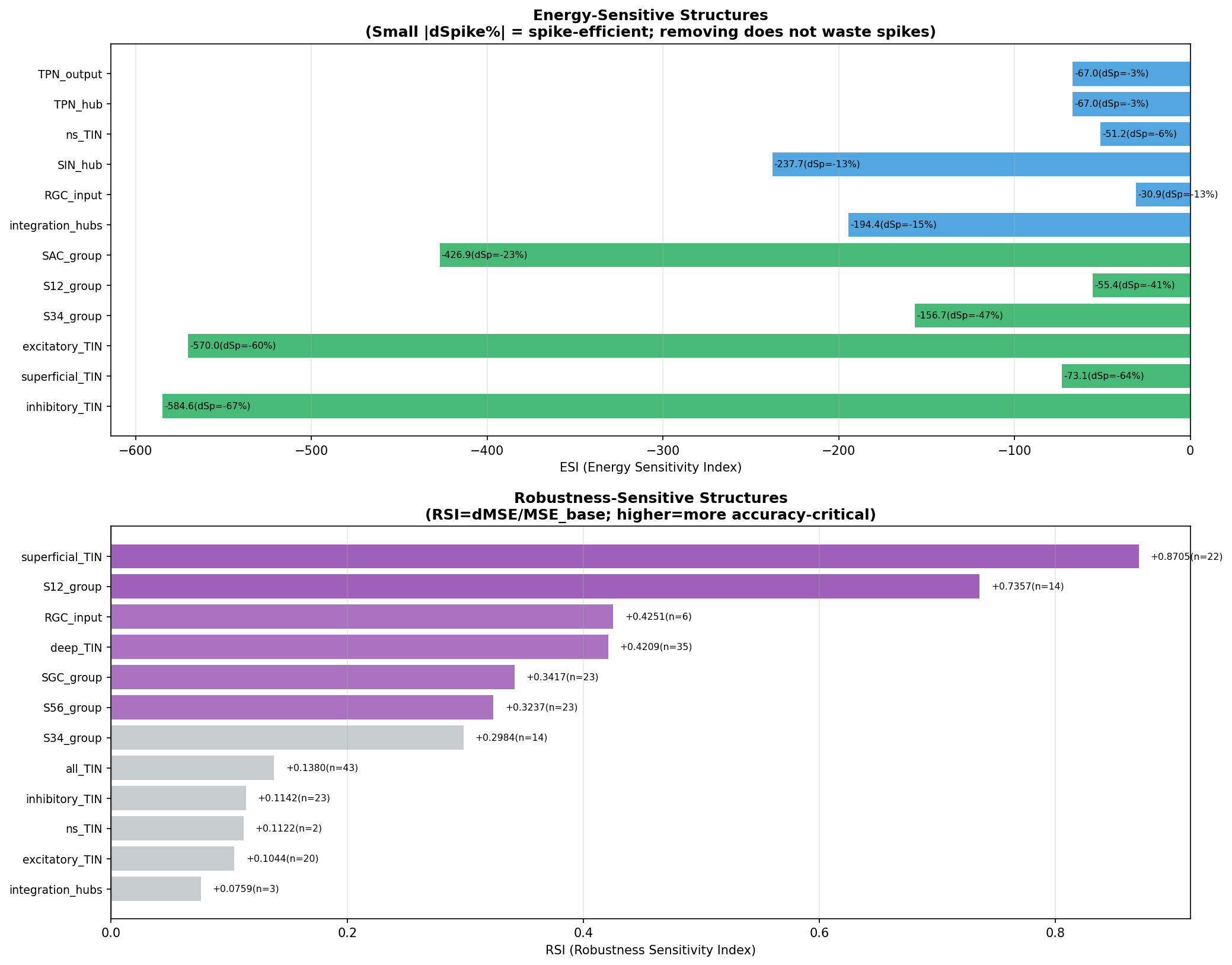}
  \caption{SNN substructure ablation results. The ESI analysis identifies \textit{ns\_TIN} as the most representative internal energy-efficient candidate after excluding the visual input group, whereas the RSI analysis identifies \textit{superficial\_TIN} as the most robustness-efficient substructure.}
  \label{fig:snn_ablation_results}
\end{figure}

The ESI ranking is summarized in Table~\ref{tab:esi_ranking} and visualized in Fig.~\ref{fig:snn_ablation_results}. 
The \textit{RGC\_input} group shows the smallest absolute ESI value, but it corresponds to the visual input port rather than an internal processing substructure. 
Because ESI is negative when ablation reduces spike count, the ranking is interpreted by $|\mathrm{ESI}|$ among ablations that increase prediction error. 
After excluding this input group, \textit{ns\_TIN} is the most representative internal energy-efficient substructure candidate. 
The \textit{ns\_TIN} substructure contains two computational nodes. 
After its ablation, the total spike count decreases by only 5.8\%, whereas the MSE increases by 0.1046. 
This indicates that \textit{ns\_TIN} has a low energy footprint, estimated by spike count, but a measurable effect on prediction performance, consistent with its interpretation as a sparse internal information gate.

\begin{table}[!htbp]
  \centering
  \caption{ESI ranking of ablated substructures. Values are reported as $100 \times \mathrm{ESI}$ for readability; because all listed ablations reduce spike count, smaller absolute values indicate a larger error cost per unit energy reduction estimated by spike count.}
  \label{tab:esi_ranking}
  \begin{tabular}{lc}
    \toprule
    Substructure & $100 \times \mathrm{ESI}$ \\
    \midrule
    \textit{RGC\_input} & -30.88 \\
    \textit{ns\_TIN} & -51.18 \\
    \textit{S12\_group} & -55.42 \\
    \textit{TPN\_output} & -66.98 \\
    \textit{superficial\_TIN} & -73.11 \\
    \bottomrule
  \end{tabular}
\end{table}

The RSI ranking is summarized in Table~\ref{tab:rsi_ranking} and visualized in Fig.~\ref{fig:snn_ablation_results}. 
The \textit{superficial\_TIN} substructure achieves the highest RSI value of 0.8705. 
Its ablation increases the MSE by 0.8112, resulting in an ablated MSE of 1.7430, while the total spike count decreases by 63.7\%. 
This indicates that \textit{superficial\_TIN} is not merely an activity-generating component, but a structure whose removal strongly disrupts system-level performance.

\begin{table}[!htbp]
  \centering
  \caption{RSI ranking of ablated substructures. RSI is the normalized MSE increase after removal and is used here as an ablation-based robustness-sensitivity proxy.}
  \label{tab:rsi_ranking}
  \begin{tabular}{lc}
    \toprule
    Substructure & RSI \\
    \midrule
    \textit{superficial\_TIN} & 0.8705 \\
    \textit{S12\_group} & 0.7357 \\
    \textit{RGC\_input} & 0.4251 \\
    \textit{deep\_TIN} & 0.4209 \\
    \textit{SGC\_group} & 0.3417 \\
    \bottomrule
  \end{tabular}
\end{table}

Together, the ESI and RSI results identify two specific sensitive substructures on different computational axes. 
\textit{ns\_TIN} is consistent with an energy-efficient bottleneck role, whereas \textit{superficial\_TIN} is better characterized as a robustness-efficient stabilizing candidate. 
Thus, the two sensitive structures are not simply different in overall importance; they occupy different computational axes.

\begin{table}[!htbp]
  \centering
  \caption{Two identified sensitive substructures in the zebrafish mesoscopic--microscopic abstraction.}
  \label{tab:functional_dissociation}
  \begin{tabular}{lcp{0.28\linewidth}p{0.34\linewidth}}
    \toprule
    Substructure & Nodes & Main evidence & Computational interpretation \\
    \midrule
    \textit{ns\_TIN} & 2 & $100 \times \mathrm{ESI} = -51.18$ & Energy-efficient information gate \\
    \textit{superficial\_TIN} & 22 & RSI = 0.8705 & Feedback-like stabilizing candidate \\
    \bottomrule
  \end{tabular}
\end{table}

\subsection{Cross-modal ANN transfer evaluation}
\label{subsec:ann_transfer_results}

The SNN ablation results identify \textit{ns\_TIN} and \textit{superficial\_TIN} as two specific prototypes for ANN transfer because they lie on different attribution axes. 
To test whether these roles remain meaningful outside the original spiking model, we evaluated two ResNet18-based variants on CIFAR-10. 
The \textit{ns\_TIN}-inspired variant was tested under inference-budget reduction, while the \textit{superficial\_TIN}-inspired variant was tested under Gaussian noise corruption.

\begin{figure}[!htbp]
  \centering
  \includegraphics[width=0.95\linewidth]{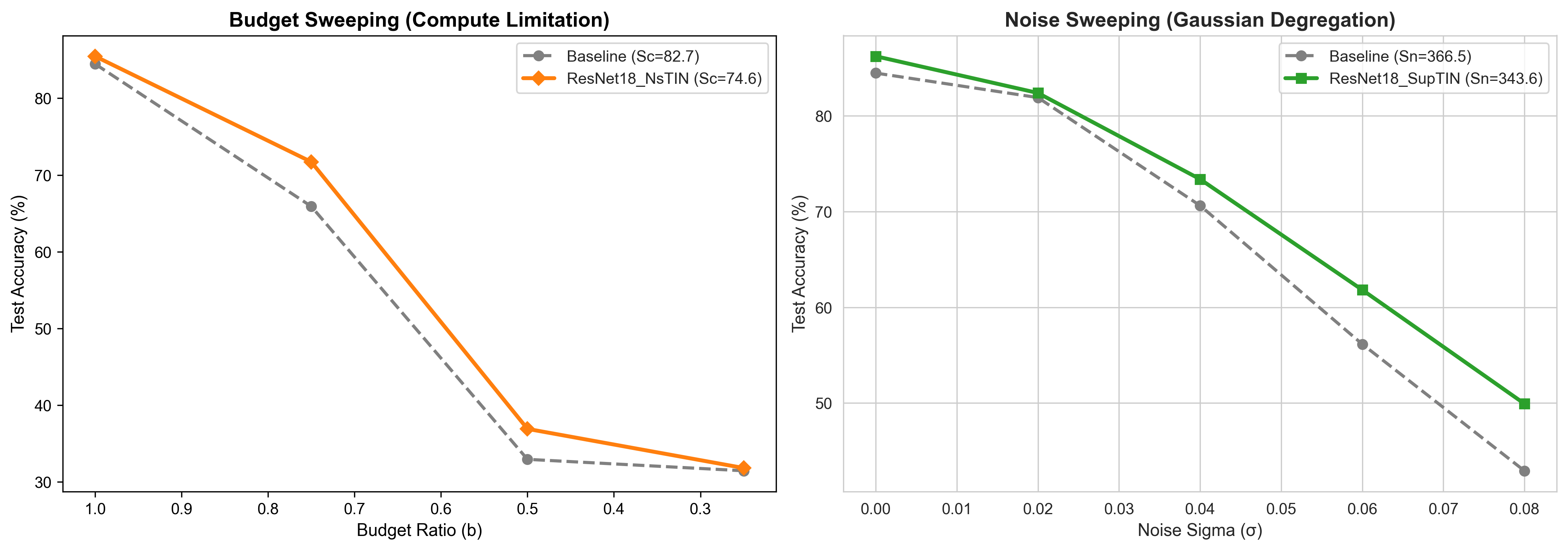}
  \caption{Cross-modal ANN transfer results. \textit{ResNet18WithNsTIN} shows slower performance degradation under inference-budget reduction, whereas \textit{ResNet18WithSuperficialTIN} maintains higher accuracy under Gaussian noise corruption.}
  \label{fig:ann_validation_results}
\end{figure}

As shown in Table~\ref{tab:budget_results} and Fig.~\ref{fig:ann_validation_results}, \textit{ResNet18WithNsTIN} achieves higher accuracy than the baseline under all tested budget ratios. 
At the full budget ratio, it achieves 85.48\% accuracy, compared with 84.48\% for the baseline. 
When the budget ratio is reduced to 0.75, the baseline accuracy drops to 65.91\%, whereas \textit{ResNet18WithNsTIN} maintains 71.71\%. 
The budget degradation score decreases from 82.70 for the baseline to 74.60 for \textit{ResNet18WithNsTIN}, indicating slower performance degradation under reduced inference capacity.

\begin{table}[!htbp]
  \centering
  \caption{Budget sweeping results on CIFAR-10.}
  \label{tab:budget_results}
  \begin{tabular}{ccc}
    \toprule
    Budget ratio & BaselineResNet18 & \textit{ResNet18WithNsTIN} \\
    \midrule
    1.00 & 84.48\% & 85.48\% \\
    0.75 & 65.91\% & 71.71\% \\
    0.50 & 32.93\% & 36.90\% \\
    0.25 & 31.44\% & 31.80\% \\
    \bottomrule
  \end{tabular}
\end{table}

As shown in Table~\ref{tab:noise_results} and Fig.~\ref{fig:ann_validation_results}, \textit{ResNet18WithSuperficialTIN} achieves higher accuracy than the baseline under all tested noise levels. 
At $\sigma=0.00$, it achieves 86.24\% accuracy, compared with 84.48\% for the baseline. 
At $\sigma=0.06$, the baseline accuracy decreases to 56.15\%, whereas \textit{ResNet18WithSuperficialTIN} maintains 61.82\%. 
At $\sigma=0.08$, the proposed model maintains 49.92\% accuracy, compared with 42.91\% for the baseline. 
The noise degradation score decreases from 366.51 for the baseline to 343.56 for \textit{ResNet18WithSuperficialTIN}, indicating improved resistance to input corruption.

\begin{table}[!htbp]
  \centering
  \caption{Noise sweeping results on CIFAR-10.}
  \label{tab:noise_results}
  \begin{tabular}{ccc}
    \toprule
    Noise $\sigma$ & BaselineResNet18 & \textit{ResNet18WithSuperficialTIN} \\
    \midrule
    0.00 & 84.48\% & 86.24\% \\
    0.02 & 81.92\% & 82.40\% \\
    0.04 & 70.63\% & 73.39\% \\
    0.06 & 56.15\% & 61.82\% \\
    0.08 & 42.91\% & 49.92\% \\
    \bottomrule
  \end{tabular}
\end{table}

These ANN transfer results are consistent with the SNN attribution results. 
The \textit{ns\_TIN}-inspired module improves performance preservation under computation-budget reduction, whereas the \textit{superficial\_TIN}-inspired module improves robustness under the tested Gaussian noise corruption.

\subsection{Summary of results}
\label{subsec:results_summary}

The results support a dual-axis interpretation of zebrafish brain substructures in the proposed mesoscopic--microscopic abstraction. 
The reconstructed graph shows a moderately dense and non-uniform topology and supports retinotectal signal propagation in dynamic simulation. 
The SNN baseline provides a nonlinear perturbation testbed rather than a strong absolute forecasting model. 
ESI and RSI jointly identify two specific sensitive substructures: \textit{ns\_TIN} and \textit{superficial\_TIN}. 
The ANN transfer experiments further show that both attributed functions remain meaningful under the tested CIFAR-10 budget and Gaussian-noise settings.

\section{Discussion}
\label{sec:discussion}

\subsection{Biological interpretation}
\label{subsec:bio_interpretation}

The ablation results suggest that different zebrafish brain substructures in the visual--motor abstraction may contribute to computation through distinct model-level functions. 
The nsTIN substructure resembles an information bottleneck. 
It contains only a small number of neurons, has a low estimated energy contribution, and yet its removal causes a measurable increase in prediction error. 
This suggests that nsTIN may support sparse selection, redundancy suppression, or gain modulation within the tectal circuit.

The superficialTIN substructure, in contrast, resembles a robustness-efficient stabilizing structure. 
It contains a larger superficial population, is associated with feedback-like coupling in the reconstructed graph, and its ablation causes the strongest error increase. 
This suggests, but does not prove, that superficialTIN may contribute to closed-loop correction and dynamic stabilization.

\subsection{Computational interpretation}
\label{subsec:computational_interpretation}

The biological findings can be translated into artificial neural network design principles. 
The nsTIN substructure corresponds to a sparse adaptive gate, which can regulate feature flow under limited computational budgets. 
The superficialTIN substructure corresponds to a recurrent feedback-like stabilizer, which can improve robustness under the tested noisy input conditions. 
The ESI provides an energy-efficiency attribution measure, whereas the RSI provides an ablation-based robustness-sensitivity measure.

\begin{table}[t]
  \centering
  \caption{Computational translation of biological findings.}
  \label{tab:bio_ai_translation}
  \begin{tabular}{ll}
    \toprule
    Biological finding & Computational translation \\
    \midrule
    nsTIN & Sparse adaptive gate \\
    superficialTIN & Recurrent feedback-like stabilizer \\
    ESI & Energy-efficiency attribution \\
    RSI & Robustness-efficiency attribution \\
    Visual--motor brain abstraction & Structural prior for neural architecture design \\
    \bottomrule
  \end{tabular}
\end{table}

\subsection{Why dual-axis attribution is necessary}
\label{subsec:dual_axis_necessity}

A single ablation metric is insufficient for interpreting biological and neuromorphic systems. 
If only MSE variation is considered, the energy cost of neural activity is ignored. 
If only spike count is considered, the functional contribution to task performance is ignored. 
The ESI captures low-energy and high-contribution substructures, whereas the RSI captures substructures whose removal causes large system-level degradation.

The dual-axis framework therefore separates energy-efficient gates from robustness-efficient stabilizers. 
This distinction is important because efficient computation and robust computation may rely on different biological structures.

\subsection{Implications for neural architecture design}
\label{subsec:architecture_implications}

The results suggest that biological brain abstractions may provide not only inspiration at the level of general motifs, but also measurable functional priors for designing artificial neural architectures. 
By first attributing the computational role of a biological substructure through SNN ablation and then transferring the corresponding principle into an ANN module, the proposed framework provides a more interpretable route for bio-inspired architecture design.

\subsection{Limitations}
\label{subsec:limitations}

Several limitations should be noted. 
First, the reconstructed graph is a category-level abstraction of zebrafish visual--motor brain connectivity rather than a complete neuron-level whole-brain connectome, and the directed edge values should be interpreted as connection-probability priors. 
Thus, the term whole-brain abstraction refers to an atlas-aligned visual--motor brain model with mesoscopic and microscopic levels, not an exhaustive model of every zebrafish brain region. 
Second, the Lorenz prediction task provides a nonlinear perturbation testbed, not a biological visual-behavior task. 
Third, spike count is used as a neural-activity proxy rather than a direct hardware energy measurement, and RSI is an ablation-based robustness-sensitivity score rather than a direct robustness metric. 
Finally, the ANN transfer experiments are limited to CIFAR-10, inference-budget reduction through residual-branch skipping, and Gaussian noise corruption; broader corruption types and parameter-matched baselines are needed before making claims of general architectural superiority.

\section{Conclusion}
\label{sec:conclusion}

This study proposed a closed-loop neurocomputing framework that connects zebrafish mesoscopic--microscopic brain abstraction, SNN-based functional attribution, and ANN-based cross-modal transfer. 
A $52 \times 52$ zebrafish visual--motor brain graph was constructed from biological connection priors, and its dynamic feasibility was examined using BrainPy-based simulations.

Through dual-axis SNN ablation, we identified two specific sensitive substructures on opposite attribution axes: nsTIN as an energy-efficient bottleneck candidate and superficialTIN as a robustness-efficient stabilization candidate.

The ANN transfer experiments further demonstrate that these biological principles can be mapped into artificial visual models. 
The nsTIN-inspired module improves performance under inference-budget reduction, and the superficialTIN-inspired module improves robustness under Gaussian noise corruption. 
These findings suggest that biological brain abstractions can serve as quantitative structural priors for building energy-efficient, robustness-efficient, and interpretable neural computing systems.

\bibliographystyle{cas-model2-names}
\bibliography{zarafish}

@article{baier2024visual,
  title = {The {{Visual Systems}} of {{Zebrafish}}},
  author = {Baier, Herwig and Scott, Ethan K.},
  year = 2024,
  month = aug,
  journal = {Annual Review of Neuroscience},
  volume = {47},
  number = {1},
  pages = {255--276},
  issn = {0147-006X, 1545-4126},
  doi = {10.1146/annurev-neuro-111020-104854},
  urldate = {2026-05-08},
  copyright = {http://creativecommons.org/licenses/by/4.0/},
  langid = {english}
}

@article{barker2015sensorimotor,
  title = {Sensorimotor {{Decision Making}} in the {{Zebrafish Tectum}}},
  author = {Barker, Alison J. and Baier, Herwig},
  year = 2015,
  month = nov,
  journal = {Current Biology},
  volume = {25},
  number = {21},
  pages = {2804--2814},
  issn = {09609822},
  doi = {10.1016/j.cub.2015.09.055},
  urldate = {2026-05-08},
  langid = {english}
}

@inproceedings{bau2017network,
  title = {Network {{Dissection}}: {{Quantifying Interpretability}} of {{Deep Visual Representations}}},
  shorttitle = {Network {{Dissection}}},
  booktitle = {2017 {{IEEE Conference}} on {{Computer Vision}} and {{Pattern Recognition}} ({{CVPR}})},
  author = {Bau, David and Zhou, Bolei and Khosla, Aditya and Oliva, Aude and Torralba, Antonio},
  year = 2017,
  month = jul,
  pages = {3319--3327},
  publisher = {IEEE},
  address = {Honolulu, HI},
  doi = {10.1109/CVPR.2017.354},
  urldate = {2026-05-08},
  isbn = {978-1-5386-0457-1}
}

@article{bianco2011prey,
  title = {Prey {{Capture Behavior Evoked}} by {{Simple Visual Stimuli}} in {{Larval Zebrafish}}},
  author = {Bianco, Isaac H. and Kampff, Adam R. and Engert, Florian},
  year = 2011,
  journal = {Frontiers in Systems Neuroscience},
  volume = {5},
  issn = {1662-5137},
  doi = {10.3389/fnsys.2011.00101},
  urldate = {2026-05-08}
}

@article{davies2018loihi,
  title = {Loihi: {{A Neuromorphic Manycore Processor}} with {{On-Chip Learning}}},
  shorttitle = {Loihi},
  author = {Davies, Mike and Srinivasa, Narayan and Lin, Tsung-Han and Chinya, Gautham and Cao, Yongqiang and Choday, Sri Harsha and Dimou, Georgios and Joshi, Prasad and Imam, Nabil and Jain, Shweta and Liao, Yuyun and Lin, Chit-Kwan and Lines, Andrew and Liu, Ruokun and Mathaikutty, Deepak and McCoy, Steven and Paul, Arnab and Tse, Jonathan and Venkataramanan, Guruguhanathan and Weng, Yi-Hsin and Wild, Andreas and Yang, Yoonseok and Wang, Hong},
  year = 2018,
  month = jan,
  journal = {IEEE Micro},
  volume = {38},
  number = {1},
  pages = {82--99},
  issn = {0272-1732, 1937-4143},
  doi = {10.1109/MM.2018.112130359},
  urldate = {2026-05-08},
  copyright = {https://ieeexplore.ieee.org/Xplorehelp/downloads/license-information/IEEE.html}
}

@article{delbene2010filtering,
  title = {Filtering of {{Visual Information}} in the {{Tectum}} by an {{Identified Neural Circuit}}},
  author = {Del Bene, Filippo and Wyart, Claire and Robles, Estuardo and Tran, Amanda and Looger, Loren and Scott, Ethan K. and Isacoff, Ehud Y. and Baier, Herwig},
  year = 2010,
  month = oct,
  journal = {Science},
  volume = {330},
  number = {6004},
  pages = {669--673},
  issn = {0036-8075, 1095-9203},
  doi = {10.1126/science.1192949},
  urldate = {2026-05-08},
  langid = {english}
}

@article{demarco2020neuron,
  title = {Neuron Types in the Zebrafish Optic Tectum Labeled by an {\emph{Id2b}} Transgene},
  author = {DeMarco, Elisabeth and Xu, Nina and Baier, Herwig and Robles, Estuardo},
  year = 2020,
  month = may,
  journal = {Journal of Comparative Neurology},
  volume = {528},
  number = {7},
  pages = {1173--1188},
  issn = {0021-9967, 1096-9861},
  doi = {10.1002/cne.24815},
  urldate = {2026-05-08},
  langid = {english}
}

@article{forster2020retinotectal,
  title = {Retinotectal Circuitry of Larval Zebrafish Is Adapted to Detection and Pursuit of Prey},
  author = {F{\"o}rster, Dominique and Helmbrecht, Thomas O and Mearns, Duncan S and Jordan, Linda and Mokayes, Nouwar and Baier, Herwig},
  year = 2020,
  month = oct,
  journal = {eLife},
  volume = {9},
  pages = {e58596},
  issn = {2050-084X},
  doi = {10.7554/eLife.58596},
  urldate = {2026-05-08},
  langid = {english}
}

@article{gabriel2012layerspecific,
  title = {Layer-{{Specific Targeting}} of {{Direction-Selective Neurons}} in the {{Zebrafish Optic Tectum}}},
  author = {Gabriel, Jens~P. and Trivedi, Chintan~A. and Maurer, Colette~M. and Ryu, Soojin and Bollmann, Johann~H.},
  year = 2012,
  month = dec,
  journal = {Neuron},
  volume = {76},
  number = {6},
  pages = {1147--1160},
  issn = {08966273},
  doi = {10.1016/j.neuron.2012.12.003},
  urldate = {2026-05-08},
  langid = {english}
}

@article{gahtan2005visual,
  title = {Visual {{Prey Capture}} in {{Larval Zebrafish Is Controlled}} by {{Identified Reticulospinal Neurons Downstream}} of the {{Tectum}}},
  author = {Gahtan, Ethan and Tanger, Paul and Baier, Herwig},
  year = 2005,
  month = oct,
  journal = {The Journal of Neuroscience},
  volume = {25},
  number = {40},
  pages = {9294--9303},
  issn = {0270-6474, 1529-2401},
  doi = {10.1523/JNEUROSCI.2678-05.2005},
  urldate = {2026-05-08},
  copyright = {https://creativecommons.org/licenses/by-nc-sa/4.0/},
  langid = {english}
}

@misc{geirhos2018imagenettrained,
  title = {{{ImageNet-trained CNNs}} Are Biased towards Texture; Increasing Shape Bias Improves Accuracy and Robustness},
  author = {Geirhos, Robert and Rubisch, Patricia and Michaelis, Claudio and Bethge, Matthias and Wichmann, Felix A. and Brendel, Wieland},
  year = 2018,
  publisher = {arXiv},
  doi = {10.48550/ARXIV.1811.12231},
  urldate = {2026-05-08},
  copyright = {arXiv.org perpetual, non-exclusive license}
}

@misc{goodfellow2014explaining,
  title = {Explaining and {{Harnessing Adversarial Examples}}},
  author = {Goodfellow, Ian J. and Shlens, Jonathon and Szegedy, Christian},
  year = 2014,
  publisher = {arXiv},
  doi = {10.48550/ARXIV.1412.6572},
  urldate = {2026-05-08},
  copyright = {arXiv.org perpetual, non-exclusive license}
}

@inproceedings{he2016deep,
  title = {Deep {{Residual Learning}} for {{Image Recognition}}},
  booktitle = {2016 {{IEEE Conference}} on {{Computer Vision}} and {{Pattern Recognition}} ({{CVPR}})},
  author = {He, Kaiming and Zhang, Xiangyu and Ren, Shaoqing and Sun, Jian},
  year = 2016,
  month = jun,
  pages = {770--778},
  publisher = {IEEE},
  address = {Las Vegas, NV, USA},
  doi = {10.1109/CVPR.2016.90},
  urldate = {2026-05-08},
  isbn = {978-1-4673-8851-1}
}

@misc{hendrycks2019benchmarking,
  title = {Benchmarking {{Neural Network Robustness}} to {{Common Corruptions}} and {{Perturbations}}},
  author = {Hendrycks, Dan and Dietterich, Thomas},
  year = 2019,
  publisher = {arXiv},
  doi = {10.48550/ARXIV.1903.12261},
  urldate = {2026-05-08},
  copyright = {arXiv.org perpetual, non-exclusive license}
}

@article{hochreiter1997long,
  title = {Long {{Short-Term Memory}}},
  author = {Hochreiter, Sepp and Schmidhuber, J{\"u}rgen},
  year = 1997,
  month = nov,
  journal = {Neural Computation},
  volume = {9},
  number = {8},
  pages = {1735--1780},
  issn = {0899-7667, 1530-888X},
  doi = {10.1162/neco.1997.9.8.1735},
  urldate = {2026-05-08},
  langid = {english}
}

@article{jonas2017could,
  title = {Could a {{Neuroscientist Understand}} a {{Microprocessor}}?},
  author = {Jonas, Eric and Kording, Konrad Paul},
  editor = {Diedrichsen, J{\"o}rn},
  year = 2017,
  month = jan,
  journal = {PLOS Computational Biology},
  volume = {13},
  number = {1},
  pages = {e1005268},
  issn = {1553-7358},
  doi = {10.1371/journal.pcbi.1005268},
  urldate = {2026-05-08},
  langid = {english}
}

@article{kunst2019cellularresolution,
  title = {A {{Cellular-Resolution Atlas}} of the {{Larval Zebrafish Brain}}},
  author = {Kunst, Michael and Laurell, Eva and Mokayes, Nouwar and Kramer, Anna and Kubo, Fumi and Fernandes, Ant{\'o}nio M. and F{\"o}rster, Dominique and Dal Maschio, Marco and Baier, Herwig},
  year = 2019,
  month = jul,
  journal = {Neuron},
  volume = {103},
  number = {1},
  pages = {21-38.e5},
  issn = {08966273},
  doi = {10.1016/j.neuron.2019.04.034},
  urldate = {2026-05-08},
  langid = {english}
}

@article{maass1997networks,
  title = {Networks of Spiking Neurons: {{The}} Third Generation of Neural Network Models},
  shorttitle = {Networks of Spiking Neurons},
  author = {Maass, Wolfgang},
  year = 1997,
  month = dec,
  journal = {Neural Networks},
  volume = {10},
  number = {9},
  pages = {1659--1671},
  issn = {08936080},
  doi = {10.1016/S0893-6080(97)00011-7},
  urldate = {2026-05-08},
  copyright = {https://www.elsevier.com/tdm/userlicense/1.0/},
  langid = {english}
}

@article{merolla2014million,
  title = {A Million Spiking-Neuron Integrated Circuit with a Scalable Communication Network and Interface},
  author = {Merolla, Paul A. and Arthur, John V. and {Alvarez-Icaza}, Rodrigo and Cassidy, Andrew S. and Sawada, Jun and Akopyan, Filipp and Jackson, Bryan L. and Imam, Nabil and Guo, Chen and Nakamura, Yutaka and Brezzo, Bernard and Vo, Ivan and Esser, Steven K. and Appuswamy, Rathinakumar and Taba, Brian and Amir, Arnon and Flickner, Myron D. and Risk, William P. and Manohar, Rajit and Modha, Dharmendra S.},
  year = 2014,
  month = aug,
  journal = {Science},
  volume = {345},
  number = {6197},
  pages = {668--673},
  issn = {0036-8075, 1095-9203},
  doi = {10.1126/science.1254642},
  urldate = {2026-05-08},
  langid = {english}
}

@article{neftci2019surrogate,
  title = {Surrogate {{Gradient Learning}} in {{Spiking Neural Networks}}: {{Bringing}} the {{Power}} of {{Gradient-Based Optimization}} to {{Spiking Neural Networks}}},
  shorttitle = {Surrogate {{Gradient Learning}} in {{Spiking Neural Networks}}},
  author = {Neftci, Emre O. and Mostafa, Hesham and Zenke, Friedemann},
  year = 2019,
  month = nov,
  journal = {IEEE Signal Processing Magazine},
  volume = {36},
  number = {6},
  pages = {51--63},
  issn = {1053-5888, 1558-0792},
  doi = {10.1109/MSP.2019.2931595},
  urldate = {2026-05-08},
  copyright = {https://ieeexplore.ieee.org/Xplorehelp/downloads/license-information/IEEE.html}
}

@article{niell2005functional,
  title = {Functional {{Imaging Reveals Rapid Development}} of {{Visual Response Properties}} in the {{Zebrafish Tectum}}},
  author = {Niell, Cristopher M. and Smith, Stephen J},
  year = 2005,
  month = mar,
  journal = {Neuron},
  volume = {45},
  number = {6},
  pages = {941--951},
  issn = {08966273},
  doi = {10.1016/j.neuron.2005.01.047},
  urldate = {2026-05-08},
  langid = {english}
}

@article{olshausen1996emergence,
  title = {Emergence of Simple-Cell Receptive Field Properties by Learning a Sparse Code for Natural Images},
  author = {Olshausen, Bruno A. and Field, David J.},
  year = 1996,
  month = jun,
  journal = {Nature},
  volume = {381},
  number = {6583},
  pages = {607--609},
  issn = {0028-0836, 1476-4687},
  doi = {10.1038/381607a0},
  urldate = {2026-05-08},
  copyright = {http://www.springer.com/tdm},
  langid = {english}
}

@article{pfeiffer2018deep,
  title = {Deep {{Learning With Spiking Neurons}}: {{Opportunities}} and {{Challenges}}},
  shorttitle = {Deep {{Learning With Spiking Neurons}}},
  author = {Pfeiffer, Michael and Pfeil, Thomas},
  year = 2018,
  month = oct,
  journal = {Frontiers in Neuroscience},
  volume = {12},
  pages = {774},
  issn = {1662-453X},
  doi = {10.3389/fnins.2018.00774},
  urldate = {2026-05-08}
}

@misc{qian2025adaptive,
  title = {An {{Adaptive Visuomotor Transformation Reservoir Embedded}} in the {{Vertebrate Brain}}},
  author = {Qian, Yu and Li, Sha and Chen, Ming-Chuan and Du, Xu-Fei and Du, Jiu-Lin},
  year = 2025,
  month = jun,
  publisher = {Neuroscience},
  doi = {10.1101/2025.06.10.658856},
  urldate = {2026-04-07},
  archiveprefix = {Neuroscience},
  copyright = {http://creativecommons.org/licenses/by/4.0/},
  langid = {english}
}

@article{robles2011characterization,
  title = {Characterization of {{Genetically Targeted Neuron Types}} in the {{Zebrafish Optic Tectum}}},
  author = {Robles, Estuardo and Smith, Stephen J. and Baier, Herwig},
  year = 2011,
  journal = {Frontiers in Neural Circuits},
  volume = {5},
  issn = {1662-5110},
  doi = {10.3389/fncir.2011.00001},
  urldate = {2026-05-08}
}

@article{roy2019spikebased,
  title = {Towards Spike-Based Machine Intelligence with Neuromorphic Computing},
  author = {Roy, Kaushik and Jaiswal, Akhilesh and Panda, Priyadarshini},
  year = 2019,
  month = nov,
  journal = {Nature},
  volume = {575},
  number = {7784},
  pages = {607--617},
  issn = {0028-0836, 1476-4687},
  doi = {10.1038/s41586-019-1677-2},
  urldate = {2026-05-08},
  langid = {english}
}

@article{rueckauer2017conversion,
  title = {Conversion of {{Continuous-Valued Deep Networks}} to {{Efficient Event-Driven Networks}} for {{Image Classification}}},
  author = {Rueckauer, Bodo and Lungu, Iulia-Alexandra and Hu, Yuhuang and Pfeiffer, Michael and Liu, Shih-Chii},
  year = 2017,
  month = dec,
  journal = {Frontiers in Neuroscience},
  volume = {11},
  pages = {682},
  issn = {1662-453X},
  doi = {10.3389/fnins.2017.00682},
  urldate = {2026-05-08}
}

@article{semmelhack2014dedicated,
  title = {A Dedicated Visual Pathway for Prey Detection in Larval Zebrafish},
  author = {Semmelhack, Julia L and Donovan, Joseph C and Thiele, Tod R and Kuehn, Enrico and Laurell, Eva and Baier, Herwig},
  year = 2014,
  month = dec,
  journal = {eLife},
  volume = {3},
  pages = {e04878},
  issn = {2050-084X},
  doi = {10.7554/eLife.04878},
  urldate = {2026-05-08},
  copyright = {http://creativecommons.org/licenses/by/4.0/},
  langid = {english}
}

@article{spoerer2017recurrent,
  title = {Recurrent {{Convolutional Neural Networks}}: {{A Better Model}} of {{Biological Object Recognition}}},
  shorttitle = {Recurrent {{Convolutional Neural Networks}}},
  author = {Spoerer, Courtney J. and McClure, Patrick and Kriegeskorte, Nikolaus},
  year = 2017,
  month = sep,
  journal = {Frontiers in Psychology},
  volume = {8},
  pages = {1551},
  issn = {1664-1078},
  doi = {10.3389/fpsyg.2017.01551},
  urldate = {2026-05-08}
}

@misc{szegedy2013intriguing,
  title = {Intriguing Properties of Neural Networks},
  author = {Szegedy, Christian and Zaremba, Wojciech and Sutskever, Ilya and Bruna, Joan and Erhan, Dumitru and Goodfellow, Ian and Fergus, Rob},
  year = 2013,
  publisher = {arXiv},
  doi = {10.48550/ARXIV.1312.6199},
  urldate = {2026-05-08},
  copyright = {Creative Commons Attribution 3.0 Unported}
}

@article{tavanaei2019deep,
  title = {Deep Learning in Spiking Neural Networks},
  author = {Tavanaei, Amirhossein and Ghodrati, Masoud and Kheradpisheh, Saeed Reza and Masquelier, Timoth{\'e}e and Maida, Anthony},
  year = 2019,
  month = mar,
  journal = {Neural Networks},
  volume = {111},
  pages = {47--63},
  issn = {08936080},
  doi = {10.1016/j.neunet.2018.12.002},
  urldate = {2026-05-08},
  langid = {english}
}

@inproceedings{xie2019feature,
  title = {Feature {{Denoising}} for {{Improving Adversarial Robustness}}},
  booktitle = {2019 {{IEEE}}/{{CVF Conference}} on {{Computer Vision}} and {{Pattern Recognition}} ({{CVPR}})},
  author = {Xie, Cihang and Wu, Yuxin and Maaten, Laurens Van Der and Yuille, Alan L. and He, Kaiming},
  year = 2019,
  month = jun,
  pages = {501--509},
  publisher = {IEEE},
  address = {Long Beach, CA, USA},
  doi = {10.1109/CVPR.2019.00059},
  urldate = {2026-05-08},
  copyright = {https://doi.org/10.15223/policy-029},
  isbn = {978-1-7281-3293-8}
}

@article{yamins2016using,
  title = {Using Goal-Driven Deep Learning Models to Understand Sensory Cortex},
  author = {Yamins, Daniel L K and DiCarlo, James J},
  year = 2016,
  month = mar,
  journal = {Nature Neuroscience},
  volume = {19},
  number = {3},
  pages = {356--365},
  issn = {1097-6256, 1546-1726},
  doi = {10.1038/nn.4244},
  urldate = {2026-05-08},
  langid = {english}
}

@article{yin2019optic,
  title = {Optic Tectal Superficial Interneurons Detect Motion in Larval Zebrafish},
  author = {Yin, Chen and Li, Xiaoquan and Du, Jiulin},
  year = 2019,
  month = apr,
  journal = {Protein \& Cell},
  volume = {10},
  number = {4},
  pages = {238--248},
  issn = {1674-800X, 1674-8018},
  doi = {10.1007/s13238-018-0587-7},
  urldate = {2026-05-08},
  langid = {english}
}

\end{document}